\documentclass[letterpaper]{article}
\usepackage{flairs}
\usepackage{times}
\usepackage{helvet}
\usepackage{graphicx}
\usepackage{courier}
\usepackage{amsmath,amssymb,amsfonts}
\usepackage{algorithmic}

\begin{document}
%
\title{Vision-Based American Sign Language Classification Approach via Deep Learning}

\author{Nelly Elsayed$^{\dagger}$, Zag ElSayed$^{\dagger}$, Anthony S. Maida$^{\mathsection}$ \\
	$^{\dagger}$University of Cincinnati\\
	$^{\mathsection}$ University of Louisiana at Lafayette \\
}
\maketitle
\begin{abstract}
\begin{quote}
Hearing-impaired is the disability of partial or total hearing loss that causes a significant problem for communication with other people in society. American Sign Language (ASL) is one of the sign languages that most commonly used language used by Hearing impaired communities to communicate with each other. In this paper, we proposed a simple deep learning model that aims to classify the American Sign Language letters as a step in a path for removing communication barriers that are related to disabilities.
\end{quote}
\end{abstract}

\noindent

\begin{figure}
	\centering
	\includegraphics[width=7cm,height=4cm]{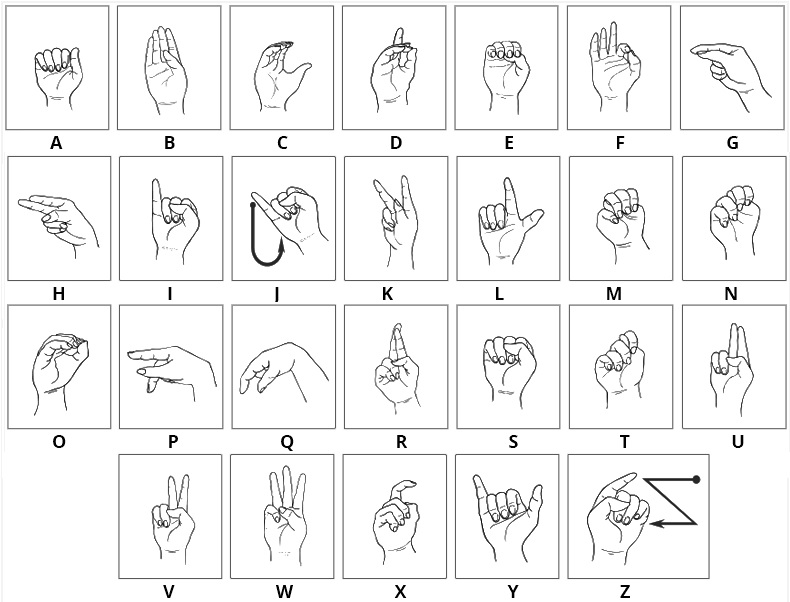}
	\caption{The American Sign Language (ASL) alphabet~\cite{alphabet}.}
	\label{alphabet_fig}
\end{figure}

Hearing-impaired is the description of a hearing disorder that is considered as any degree of hearing loss~\cite{demorest1987development}. Sign language is one of the most common communication ways that is used among hearing-impaired people~\cite{pfau2012sign}. Sign language includes gestures, body movements, and facial expressions to represent words, tone, and emotions instead of using sounds~\cite{sandler2006sign}. According to the World Federation of the Deaf (WFD), over 72 million people around the earth are deaf. In addition, there are more than 300 different sign languages that exist and are used by different deaf and hard-of-hearing people around the world~\cite{WinNT}. Sign languages do not maintain the same grammatical properties as spoken languages. Nevertheless, the sign languages maintain similar linguistic properties as the spoken languages~\cite{battison1974phonological}. The Americal Sign Language (ASL) is the most commonly used sign language in the United States and several parts of Canada~\cite{hill2018sign}. The ASL is considered to be originated in 1817 at the American School of Deaf (ASD)~\cite{bahan1996non} where the signs have been adopted from the French sign language~\cite{valli2000linguistics}. Figure~\ref{alphabet_fig} shows the ASL alphabet signs~\cite{alphabet}.
Fingers spelling is a standard system used in different sign languages to spell names, locations, words, and phrases that do not have a specific sign and also to clarify words when a specific sign was not well provided.
Interpreting sign language to a speech is essential to remove communication barriers and provide a higher quality of life for deaf and hard of hearing people worldwide. 
The ASL letters classification is a complex problem due to the large variety of different representations for the same letter due to the different physical abilities to move fingers to represent the letter and the length of the fingers. There were few attempts to solve that problem, such as~\cite{abdulhussein2020hand} that targeted the ASL letters classification by applying a deep learning model and edge detection for the hand and fingers. However, this work did not summarize prediction results on a classification problem regarding each alphabet letter. In addition, the dataset contained only 240 images where ten different samples represented each letter.~\cite{ameen2017convolutional} proposed a convolution model that attempted to classify the ASL letters. This work focused on using different types of features that were used in~\cite{rioux2014sign} combined by the convolution neural network. The model had higher accuracy than the~\cite{rioux2014sign} approaches. However, the lack of a dataset and the model implementation complexity was the major issue of this model.

\begin{figure*}
	\centering
	\includegraphics[width=12cm,height=2.5cm]{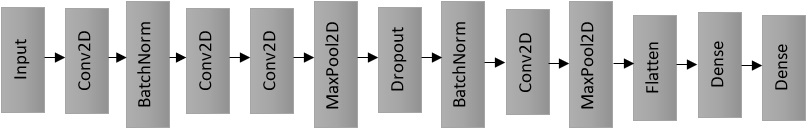}
	\caption{The proposed model for ASL lettes classification architecture.}
	\label{model_architecture}
\end{figure*}

This paper proposed an ASL classification approach via a deep convolution neural network. The proposed model can classify ASL hand postures images to their corresponding letters. In this paper, we addressed the major issues of the ALS sign classification problem. The contribution of this paper is as follows: we employed data augmentation~\cite{antoniou2017data} via multiple augmentation approaches to solving the data limitations in the ASL letters classification problem. In addition, we empirically designed a simple convolution neural network-based model that achieved a classification accuracy and can be trained rapidly compared to the other existing models. Moreover, this model does not require any data preprocessing other than image size adjustment. Furthermore, the model performs the classification without additional segmentation algorithms or transfer learning techniques. Finally, this model is employed within a system that interprets the American Sign Language letters to caption words that help readers understand the ASL speakers and remove the communication barriers.



\begin{table}
	\setlength{\tabcolsep}{2.5pt}
	\caption{The proposed model for ASL letters classification summary.} 
	\footnotesize
	\centering 
	\begin{tabular}{ l  l l} 
		\textbf{Layer} &\textbf{Output Shape}&\textbf{\# Parameters}\\
		\hline
		Input layer	&  [(None, 50, 50, 3)] & 0 \\
		Conv2D&     (None, 48, 48, 32)    &   896 \\
		Batch Normalization& (None, 48, 48, 32)    &    128  \\     
		Conv2D      &    (None, 46, 46, 64)   &   18496     \\
		Conv2D       &    (None, 44, 44, 128)    &   73856     \\
		MaxPooling2D & (None, 22, 22, 128)     &  0      \\   
		Dropout &            (None,22, 22, 128)    &   0   \\      
		Batch Normalization & (None, 22, 22, 128)  &     512    \\   
		Conv2D         &   (None, 20, 20, 256)   &      295168   \\ 
		MaxPooling2D & (None, 10, 10, 256)     &    0         \\
		Flatten &            (None, 25600)         &     0         \\
		Dense &                (None, 64)       &         1638464    \\
		Dense     &         (None, 30)    &            1950  \\
		\hline
	\end{tabular}
	\label{table_summary} 
\end{table}

\section{Proposed Model} \label{proposed_model}
The proposed model is mainly based on the convolution neural network architecture (CNN) as a robust algorithm for images classification task~\cite{lawrence1997face,howard2013some,yadav2019deep}.
The proposed model design has been selected based on empirical evaluations of different convolution layers models. The choice has been set to the lightest model design while maintaining comparable accuracy. The proposed model design consists of 13-layered architecture is shown in Figure~\ref{model_architecture}. For all the convolution layers, kernel sizes were set to 3$\times$3. The number of kernels (filters) in the convolution layers was set to 32, 64, 128, and 256. We used the \textit{glorot\_uniform}~\cite{hanin2018start,glorot2010understanding} to initialize the kernel weights and the biases were initialized by \textit{zeros}. The dropout was set to 20\%. We used the
Adam optimization function~\cite{kingma2014adam} with initial learning rate $\alpha$ = 0.01, and $\beta_{1}$ = 0.9, and $\beta_{2}$ = 0.999. The maximum pooling pool was set to 2$\times$2, the padding was set as \textit{valid}, and the strides were set to 1$\times$1. The Flatten layer is used to adjust the input data size before the fully connected dense layer. The dense layer units were set to 64. The weights of the dense layer were initialized using \textit{glorot\_uniform} function, and the ReLU function was used as the activation function as it has minimal cost compared to the other non-linear activation functions~\cite{teh2000rate,8576058}. Finally, the softmax layer was used to classify the 26 letters of the ASL and the three gesture signs: space, delete, and nothing. The proposed model summary is shown in Table~\ref{table_summary}.

\begin{figure}
	\centering
	\includegraphics[width=7cm,height=4cm]{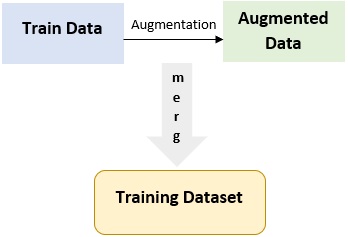}
	\caption{The dataset preparation by applying data augmentation.}
	\label{data_augmentation}
\end{figure}

\section{Data Preparation} \label{data_preparation}
Data augmentations are techniques that aim to increase the existing data by performing some different modifications on the copies of the original data. Data augmentation has been used in several data analysis and machine learning tasks where there was a lack of training data availability to train the model. In addition, data augmentation acts as a model regularizer that helps reduce and prevent the overfitting problem during the model training. Our model applied four types of data augmentation: gaussian noise~\cite{lopes2019improving}, image rotation by 90 degrees, image rotation by 30 degrees, and image rotation by -60 degrees~\cite{shorten2019survey}. Each augmentation type was applied to a randomly selected quarter of the dataset, maintaining the uniqueness of each image selection for the augmentation. Applying augmentation increased the dataset size. We used the ASL Alphabet dataset, which is available on Kaggle~\cite{alphabet_dataset}. The augmentation process is shown in Figure~\ref{data_augmentation}. The original dataset size was 87,000 images, and the dataset size was increased to 108,627 images after employing the data augmentation. Then, we performed the data normalization and image cropping to 50$\times$50 as a data preprocessing stage before the data splitting. After augmentation and data preprocessing, we split the dataset into 60\% for training, 20\% for validation, and 20\% for testing. The images in this dataset have different pixels intensity. 


\begin{figure}
	\centering
	\includegraphics[width=6cm,height=3cm]{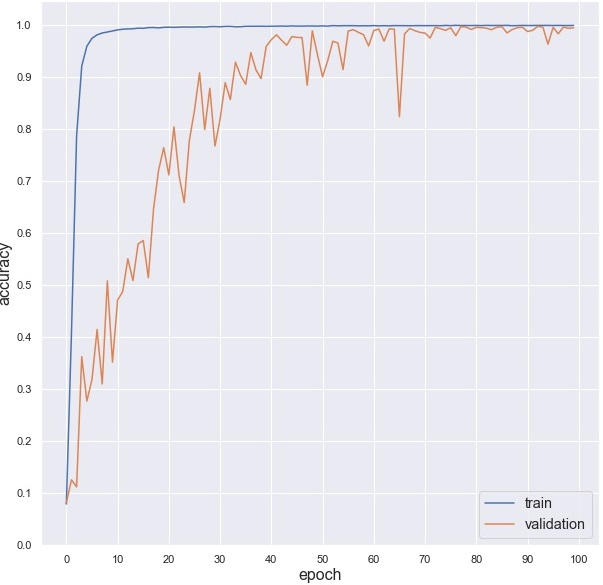}
	\caption{The training versus validation accuracy throguht the 100 epochs.}
	\label{accuracy_plot}
\end{figure}

\begin{figure}
	\centering
	\includegraphics[width=6cm,height=3cm]{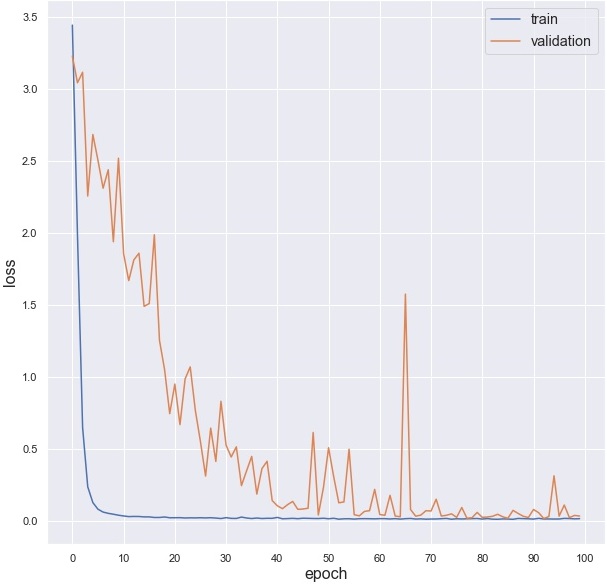}
	\caption{The training versus validation loss throguht the 100 epochs.}
	\label{loss_plot}
\end{figure}

\section{Experiments and Results} \label{results_section}

\begin{figure}
	\centering
	\includegraphics[width=8cm,height=5cm]{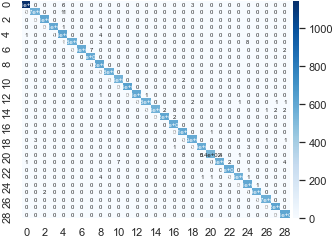}
	\caption{The confusion matrix of the poposed model testing results.}
	\label{confusion_matrix}
\end{figure}

Our experiments were performed on a Window 10 OS, Intel(R) Core(TM) i-9 CPU @ 3.00 GHz processor, and NVIDIA GeForce RTX 2080 Ti. We used Tensorflow 2.4.0, Python 3.8, and NumPy 1.19.5. The model was trained for 100 epochs. The batch size was set to 128. The RMSProp has been used as the model optimization function~\cite{hinton2012neural}. The loss function was set to the categorical crossentropy function~\cite{ketkar2017introduction}. The training versus validation accuracy is shown in Figure~\ref{accuracy_plot}. The loss of training versus validation is shown in Figure~\ref{loss_plot}. The empirical results of our proposed model are shown in Table~\ref{table_results}. Figure~\ref{confusion_matrix} shows the confusion matrix of the proposed model, where the numbers 0 to 29 indicate the alphabet letter starting from the letter A to letter Z, in addition to the nothing, delete, and space gesture signs.

\begin{table}
	\setlength{\tabcolsep}{2.5pt}
	\caption{The emperical result of our proposed model.} 
	\footnotesize
	\centering 
	\begin{tabular}{ l  l } 
		\textbf{Comparison} &\textbf{Value}\\
		\hline
		Train Accuracy	& 99.949\% \\
		Test Accuracy&    99.889\%\\
		Train Time& 55.77(min)\\
		Test Time& 0.2363(sec)\\
		Precision& 0.9849\\
		Recall& 0.9924\\
		F1-score&0.9925\\ 
        Total Parameters& 2,029,470\\
        Trainable Parameters&2,029,150\\
        Non-trainable Parameters& 320\\
		\hline
	\end{tabular}
	\label{table_results} 
\end{table}

Table~\ref{comparison_table} shows the comparison results between our proposed model with other research works that address the sign language gesture classification problem using different gesture-based datasets. 

\begin{table*}
	\setlength{\tabcolsep}{2.5pt}
	
	\caption{A comparison between different gestures classification models from the method, dataset, number of signs, and accuracy aspects.} 
	\footnotesize
	\centering 
	\begin{tabular}{ l  l l l  l} 
		\textbf{Author} &\textbf{Method}&\textbf{Dataset}&\textbf{\#Signs} &\textbf{Accuracy(\%)}\\
		\hline	
		\cite{elpeltagy2018multi}&  $\mathrm{(HOG–PCA)+(COV3DJ)+(CCA)}$& ChaLearn& 20 & 83.12\% \\
		\cite{ansari2016nearest}& $\mathrm{ANNs, 100-neurons}$&Indian-reduced&20 &37.27\%\\
		\cite{ansari2016nearest}& $\mathrm{ANNs, 400-neurons}$&Indian-reduced&2 &90.979\%\\
		 \cite{elpeltagy2018multi}& $\mathrm{(HOG–PCA)+(CCA)}$& Indian&140 &60.40\%\\
		 \cite{ansari2016nearest}& $\mathrm{SIFT}$&Indian&140 &49.07\%\\		 
		 \cite{quinn2019british}&$\mathrm{HOG-SVM-RBF}$&British&26& 99.00\%\\
		 \cite{quinn2019british}&$\mathrm{HOG-SVM-Linear}$&British&26& 98.89\%\\
		 \cite{nagarajan2013static}&$\mathrm{EOH + SVM}$&British&26& 93.75\%\\
		 \cite{barkoky2011static}&$\mathrm{ANN-Backpropagation}$&Digits&10& 96.62\%\\
		 \cite{barbhuiya2021cnn}&$\mathrm{VGG16+SVM}$&ASL&36& 99.76\%\\
		 \cite{barbhuiya2021cnn}&$\mathrm{AlexNet+SVM}$&ASL&36& 99.82\%\\
		\textbf{Our} &Simple $\mathrm{CNN}$&ASL &29& \textbf{99.94}\%\\	
		\hline
	\end{tabular}
	\label{comparison_table} 
\end{table*}

\section{Conclusion and Future Work} \label{Conclusion}
The communication gap between the hearing-impaired and hearing people has been one of the significant issues in all societies for decades. The proposed model aims to reduce the misunderstanding gap between hearing-impaired and hearing people by understanding the American Sign Language (ASL) letter via classification. The proposed model achieved significantly high accuracy for the correct classification of the ASL letters.

As future work, this model is a part of a project to translate the classified letters into transcript words that can also be converted to voice. That could help eliminate the communication gap between hearing-impaired and hearing people and provide a better quality of life for deaf and hard of hearing people, which can significantly improve social communication and understanding.

\bibliographystyle{flairs}
\bibliography{references_ALS_Clasification}

\end{document}